\documentclass{article}
\usepackage{ijcai18}

\usepackage{times}
\usepackage{xcolor}
\usepackage{soul}
\usepackage[utf8]{inputenc}
\usepackage[small]{caption}

\usepackage{graphicx}
\usepackage{amsmath, amsfonts}
\usepackage{savesym}
\usepackage{multirow}
\usepackage{booktabs}
\usepackage{url}
\usepackage{color}
\usepackage{xcolor}
\usepackage{setspace}
\usepackage{bm}
\usepackage{amssymb}
\usepackage{pifont}

\newcommand{\argmax}{\mathop{\rm argmax}\limits}
\newcommand{\argmin}{\mathop{\rm argmin}\limits}

\usepackage{latexsym}

\title{Interpretable Adversarial Perturbation in Input Embedding Space for Text}

\author{
Motoki Sato$^1$\thanks{This work was conducted when the first author worked at Nara Institute of Science and Technology and RIKEN AIP.},
Jun Suzuki$^{2,4}$\thanks{His current affiliation is Tohoku University.},
Hiroyuki Shindo$^{3,4}$,
Yuji Matsumoto$^{3,4}$\\
$^1$Preferred Networks, Inc.,\\
$^2$NTT Communication Science Laboratories,\\
$^3$Nara Institute of Science and Technology,\\
$^4$RIKEN Center for Advanced Intelligence Project\\
\texttt{sato@preferred.jp}, \texttt{suzuki.jun@lab.ntt.co.jp}\\
\texttt{\{shindo, matsu\}@is.naist.jp}
}
\begin{document}

\maketitle

\begin{abstract}
 Following great success in the image processing field, the idea of adversarial training has been applied to tasks in the natural language processing (NLP) field.
 One promising approach directly applies adversarial training developed in the image processing field to the input word embedding space instead of the discrete input space of texts.
 However, this approach abandons such {\it interpretability} as generating adversarial texts to significantly improve the performance of NLP tasks.
 This paper restores interpretability to such methods by
 restricting the directions of perturbations toward the existing words in the input embedding space.
 As a result, we can straightforwardly reconstruct each input with perturbations to an actual text by considering the perturbations to be the replacement of words in the sentence while maintaining or even improving the task performance\footnote{Our code for reproducing our experiments is available at \url{https://github.com/aonotas/interpretable_adv}}.
\end{abstract}

\section{Introduction}
The existence of (small) perturbations, which induce prediction error in machine learning models, was first discovered and discussed in~\cite{Szegedy2013IntriguingPO}.
They called the perturbed inputs {\it adversarial examples}.
Such perturbations can be easily found by optimizing the input to maximize the prediction error.
After this discovery, a framework called {\it adversarial training} (AdvT) was proposed~\cite{Goodfellow2014ExplainingAH} whose basic idea was to train models that can correctly classify both the original training data and adversarial examples generated based on the training data.
Using AdvT, we can further improve the generalization performance of models.
This improvement implies that the loss function of adversarial examples works as a good {\em regularizer} during model training.
 Currently, a technique for generating adversarial examples is crucial to neural image processing for both improving the task performance and analyzing the behaviors of {\it black-box} neural models.
 %

 %%%%%%%%%%%%%%%%%%%%%%%%%%
  Unlike its great success in the image processing field,
  AdvT cannot be straightforwardly applied to tasks in the natural language processing (NLP) field.
  This is because we cannot calculate the {\it perturbed inputs} for tasks in the NLP field since the inputs consist of discrete symbols, which are not a continuous space used in image processing.
   A novel strategy was recently proposed to improve AdvT for NLP tasks~\cite{miyato2016adversarial} whose
  key strategy is simple and straightforward:
  applying AdvT to continuous word embedding space rather than the discrete input space of texts.
  Their method preserves a theoretical background developed in the image processing field and works well as a regularizer.
  In fact, this method significantly improved the task performance and achieved the state-of-the-art performance on several text classification tasks.
  Another notable merit of this method is succinct architecture.
  It only requires the gradient of the loss function to obtain adversarial perturbations (see Eq.~\ref{eq:adv_approx}).
  Note that the gradient calculation is a standard calculation procedure for updating the model parameters during training.
  We can obtain adversarial perturbations in the embedding space with a surprisingly small calculation cost without incorporating any additional sophisticated architecture.

%%%%%%%%%%%%%%%%%%%%%%%%%%%%%%
\begin{figure}[t]
 \begin{center}
  \includegraphics[width=\linewidth]{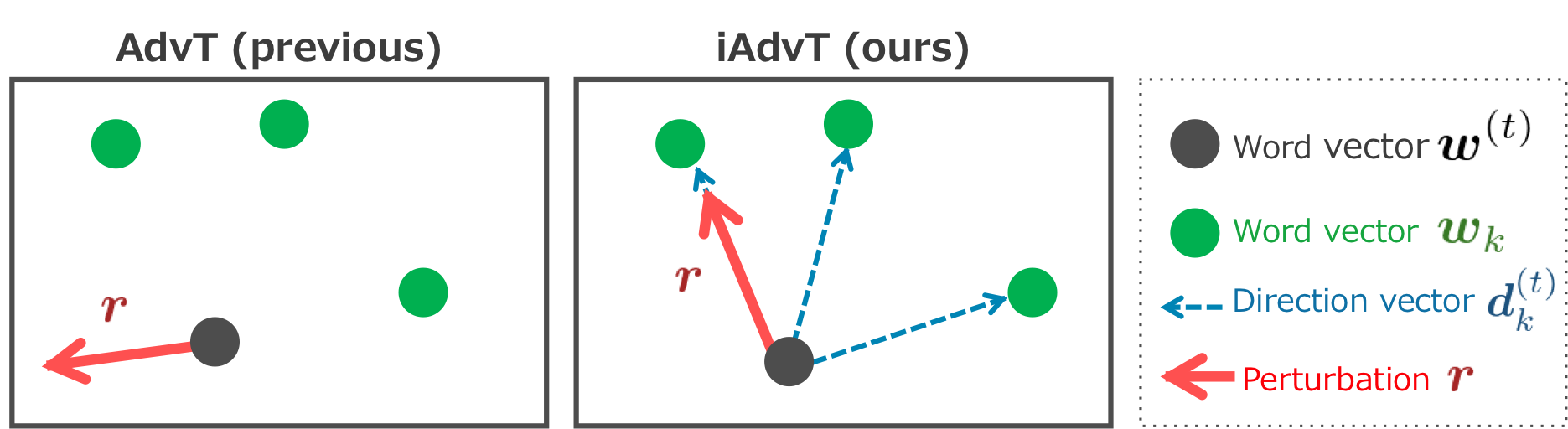}
  \caption{Intuitive sketch to explain our idea: our method (right) restricts perturbations in which words exist in the input word embedding space, whereas previous method (left) allows them to select any direction.}
  \label{fig:overview_of_our_method}
 \end{center}
\end{figure}
%%%%%%%%%%%%%%%%%%%%%%%%%%%%%%

  %
  In contrast, the main drawback of this method is that it abandons the generation of adversarial examples interpretable by people
  since how to appropriately reconstruct perturbed inputs in the input word embedding space to actual texts is not trivial.
  This implies that this approach lacks {\it interpretability}.
  In fact,
  they declared that since their training strategy is no longer intended as a defense against adversaries, they exclusively proposed it as a regularizer to stabilize the model~\cite{miyato2016adversarial}.
  It is often desirable for researchers and developers to generate adversarial examples (adversarial texts) to understand the behavior of {\em black-box} neural models.
  Therefore, a trade-off exists between well-formed and low-cost (gradient-based) approaches and the interpretability of the AdvT methods used in the NLP field.
%

%%%%%%%%%%%%%%%%%%%%%%%
 The main topic of this paper is the reduction of this trade-off gap.
 This paper restores interpretability while preserving the good ability of regularizer.
 Our main idea is to only restrict the directions of the perturbations toward the locations of existing words in the word embedding space.
 Fig.~\ref{fig:overview_of_our_method} illustrates an intuitive explanation of our idea.
 With our method, we can straightforwardly interpret each input with a perturbation as an actual sentence by considering the perturbations to be substitutions of the words in the sentence.
 To the best of our knowledge, our study is the first trial that discusses the interpretability of AdvT based on adversarial perturbation applied to tasks in the NLP field.

%%%%%%%%%%%%%%%%%%%%%%%%%%%%%%%
\section{Related Work}
Several studies have applied the ideas of AdvT to certain NLP tasks.
A method was proposed that fooled reading comprehension systems by adding sentences to the ends of paragraphs using crowdsourcing~\cite{Jia2017AdversarialEF}.
Random character swaps can break the output of neural machine translation systems ~\cite{Belinkov2017SyntheticAN,Hosseini2017DeceivingGP}, and thus they proposed AdvT methods that generate random character swaps and utilize the generated input sentences as additional training data for their models.
Moreover, a method generated a large number of input sentences by replacing a word with its synonym~\cite{samanta2017towards}

%%%%%%%%%%%%%%%%%%%%%%%%%
The primary
strategy for generating adversarial examples in the NLP field clearly differs from those developed in the image processing field, which
 are rather ad-hoc, e.g., using human knowledge~\cite{Jia2017AdversarialEF}, dictionaries~\cite{samanta2017towards},
or require such costly procedures as exhaustive searches~\cite{samanta2017towards}.
These methods are not essentially based on the previously discussed idea of perturbation that was first discussed~\cite{Szegedy2013IntriguingPO} for generating {\it adversarial examples}.

In contrast,
our baseline method~\cite{miyato2016adversarial} preserves a theoretical background developed in the image processing field.
Thus, note that the methods discussed in this paper borrow a distinct strategy from the current primal strategy taken in the NLP field as described above.

\section{Target Tasks and Baseline Models}

\begin{figure}[t]
  \begin{center}
  \includegraphics[width=\linewidth]{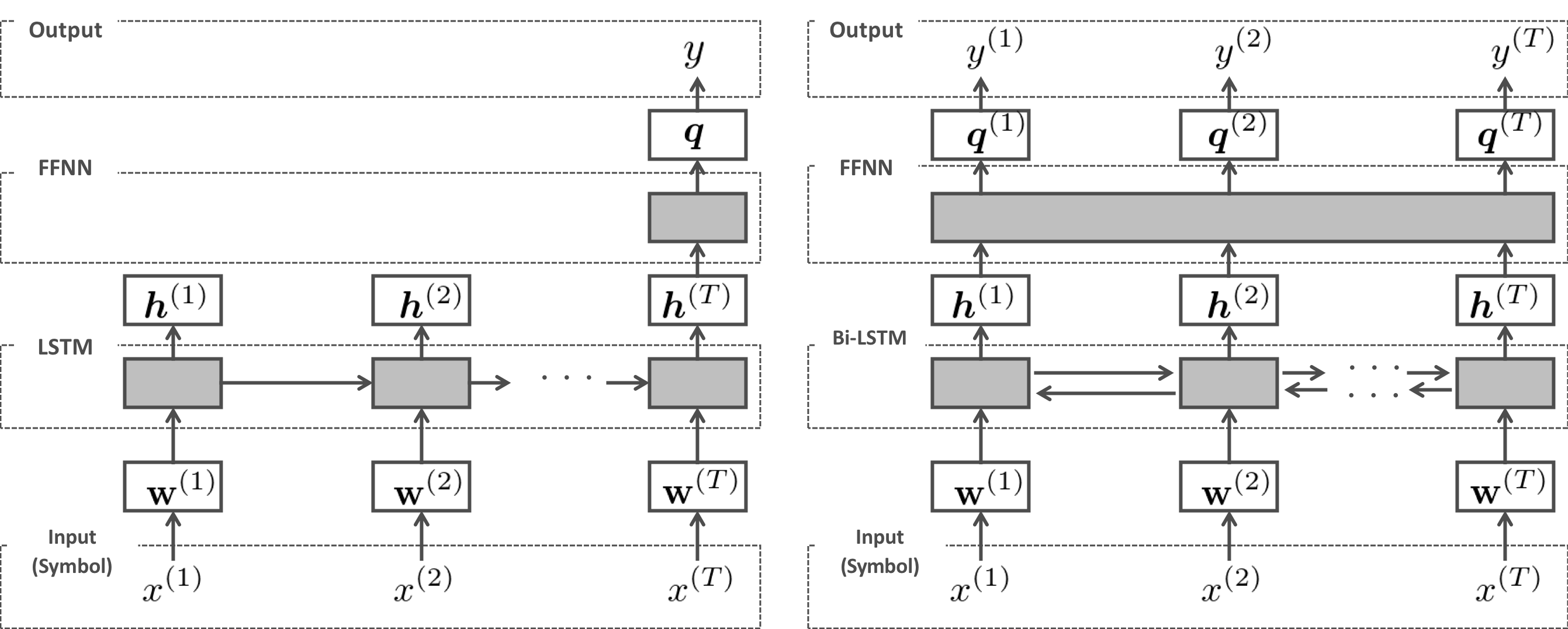}
  \caption{Overview of our baseline neural models: LSTM-based classifier for sentiment classification (left) and Bi-LSTM model for grammatical error detection (right).}
  \label{fig:lstm_baseline_overview}
  \end{center}
  \end{figure}

%%%%%%%%%%%%%%%%%%%%
This section briefly explains the formal definitions of our target tasks, text classification and sequence labeling, and the baseline neural models for modeling these tasks.
Fig.~\ref{fig:lstm_baseline_overview} shows the architecture of the baseline neural models.

\subsection{Common notation}
%%%%%%%%%%%%%%%%%%%%
Let $X$ represent an input sentence.
$\mathcal{V}$ denotes the vocabulary of the input words.
$x^{(t)}\in \mathcal{V}$ is the $t$-th word that appears in given input sentence $X$, where
$X=(x^{(1)},\dots, x^{(T)})$ if the number of words in $X$ is $T$.
Here we introduce the following short notation of sequence $(x^{(1)},\dots, x^{(T)})$ as $(x^{(t)})^T_{t=1}$.
$\mathcal{Y}$ denotes a set of output classes.
To explain the text classification and the sequence labeling tasks in a single framework,
this paper assumes that output $Y$ denotes sequence of class labels ${Y}=({y}^{(t)})^{T}_{t=1}$, where ${y}^{(t)}\in\mathcal{Y}$ for all $t$ in the case of sequence labeling,
and class label ${Y}={y}$, where ${y}\in\mathcal{Y}$ for the text classification case.

%%%%%%%%%%%%%%%%%%%
Let $\bm{w^{(t)}}$ be a word embedding vector that corresponds to $x^{(t)}$ whose dimension is $D$, where $\bm{w}^{(t)} \in {\mathbb R}^{D}$.
Thus, sequence of word embedding vectors $\tilde{X}$ that corresponds to $X$ can be written as $\tilde{X} = (\bm{w}^{(t)})^T_{t=1}$.
Then for text classification, $\tilde{y}$ denotes a corresponding class ID of $y$ in $\mathcal{Y}$.
$\tilde{y}$ always takes one integer from 1 to $|\mathcal{Y}|$, where $\tilde{y}\in \{1, \dots, |\mathcal{Y}|\}$.
$\tilde{y}^{(t)}$ also denotes a corresponding class ID of $\tilde{y}^{(t)}$ in $\mathcal{Y}$ for sequence labeling.
Finally, $\tilde{Y}$ represents $\tilde{Y}=\tilde{y}$ for text classification and $\tilde{Y}=(\tilde{y}^{(t)})^{T}_{t=1}$ for sequence labeling.

%%%%%%%%%%%%%%%%%%%
Here, without loss of generality, we formulate a text classification task or a sequence labeling task whose inputs and outputs are respectively $\tilde{X}$ and $\tilde{Y}$ instead of $X$ and $Y$.
This is because we can uniquely convert from $X$ to $\tilde{X}$ and from $\tilde{Y}$ to $Y$.
Thus, training data $\mathcal{D}$ can be represented as a set of $\tilde{X}$ and $\tilde{Y}$ pairs, namely, $\mathcal{D}=\{(\tilde{X}^{(n)},\tilde{Y}^{(n)})\}^{N}_{n=1}$, where $N$ represents the amount of training data.

%%%%%%%%%%%%%%%%%%%
%%%%%%%%%%%%%%%%%%%

\subsection{Baseline model for text classification}
%%%%%%%%%%%%%%%%%%%%%%%
We encode input $\tilde{X}$ with a recurrent neural network (RNN)-based model, which consists of an LSTM unit~\cite{hochreiter1997long}.
The (forward) LSTM unit calculates a hidden state in each step $t$ as
$ \bm{h}^{(t)} =  {\tt LSTM}\big(\bm{w}^{(t)}, \bm{h}^{(t-1)}\big)$,
where $\bm{h}^{(0)}$ is assumed to be a zero vector.
Then we model the (conditional) probability of output $\tilde{Y}$ given input $\tilde{X}$ as follows:
\begin{align}
 p(\tilde{Y}\mid \tilde{X}, \mathcal{W}) &=  \frac{\exp(q_{\tilde{y}})}{\sum^{|\mathcal{Y}|}_{m=1} \exp(q_{m})}
 ,
\end{align}
where $q_m$ is the $m$-th factor of $\bm{q}$ whose dimension is $|\mathcal{Y}|$.
$\bm{q}$ is calculated through a standard feed-forward neural network from $T$-th final hidden state $\bm{h}^{(T)}$, where
$\bm{q} = {\tt FFNN}(\bm{h}^{(T)})$.
 Here we omit an explanation of the detailed configurations of ${\tt LSTM}$ and ${\tt FFNN}$, but they will be described in our experiment section, since the selection of their configurations affects none of this paper's discussion.

\subsection{Baseline model for sequence labeling}
%%%%%%%%%%%%%%
We employ a bi-directional LSTM to encode input $\tilde{X}$.
The hidden state of each step $t$, that is, $\bm{h}^{(t)}$, can be obtained by the concatenation of two hidden states from forward and backward LSTMs:
$\bm{h}^{(t)} = {\tt concat}(\bm{h}^{(t)}_{\rm f}, \bm{h}^{(t)}_{\rm b} )$,
where $\bm{h}^{(t)}_{\rm f} =  {\tt LSTM}\big(\bm{w}^{(t)}, \bm{h}^{(t-1)}_{\rm f}\big)$,
and $\bm{h}^{(t)}_{\rm b} =  {\tt LSTM}\big(\bm{w}^{(t)}, \bm{h}^{(t+1)}_{\rm b}\big)$.
We assume that $\bm{h}^{(0)}_{\rm f}$ and $\bm{h}^{(T+1)}_{\rm b}$ are always zero vectors.
We also assume that probability $p(\tilde{Y}\mid \tilde{X}, \mathcal{W})$ can be decomposed into each step $t$.
This means that probability $p(\tilde{Y}\mid \tilde{X}, \mathcal{W})$ can be calculated:
\begin{align}
 p(\tilde{Y}\mid \tilde{X}, \mathcal{W})
 =&
 \prod^{T}_{t=1} p(\tilde{y}^{(t)} \mid \tilde{X}, \mathcal{W})
 \\
 p(\tilde{y}^{(t)} \mid \tilde{X}, \mathcal{W})
 =&
 \frac{\exp(q_{\tilde{y}^{(t)}})}{\sum^{|\mathcal{Y}|}_{m=1} \exp(q_{m})}
 ,
\end{align}
where $q^{(t)}_m$ is the $m$-th factor of $\bm{q}^{(t)}$ whose dimension is $|\mathcal{Y}|$.
$\bm{q}^{(t)}$ is calculated through a standard feed-forward neural network from $t$-th final hidden state $\bm{h}^{(t)}$:
$\bm{q}^{(t)} = {\tt FFNN}(\bm{h}^{(t)})$.

\subsection{Training}
%%%%%%%%%%%%%%%%%%%%%%%%%%%%%
For training both the text categorization and the sequence labeling, we generally find the optimal parameters of an RNN-based model that can minimize the following optimization problem:
\begin{align}
 \hat{\mathcal{W}} &=  \argmin_{\mathcal{W}} \Big\{ \mathcal{J} (\mathcal{D}, \mathcal{W}) \Big\},
\end{align}
where $\mathcal{W}$ represents the overall parameters in the RNN-based model.
$\mathcal{J} (\mathcal{D}, \mathcal{W})$ is the loss function over entire training data $\mathcal{D}$, and $\ell(\tilde{X},\tilde{Y},\mathcal{W})$ is the loss function of individual training sample $(\tilde{X},\tilde{Y})$ in $\mathcal{D}$:
\begin{align}
 \mathcal{J} (\mathcal{D}, \mathcal{W}) &= \frac{1}{|\mathcal{D}|} \sum_{(\tilde{X},\tilde{Y})\in \mathcal{D}} \ell(\tilde{X}, \tilde{Y}, \mathcal{W})
 \\
 \ell(\tilde{X},\tilde{Y},\mathcal{W}) &= - \log \big( p(\tilde{Y}\mid \tilde{X}, \mathcal{W}) \big)
 .
\end{align}

%%%%%%%%%%%%%%%%%%%%%%%%%%%%%%%%%%%%%%%%%%%%%%%%%%%%%%
\section{Adversarial Training in Embedding Space}
\begin{figure}[t]
\begin{center}
\includegraphics[width=\linewidth]{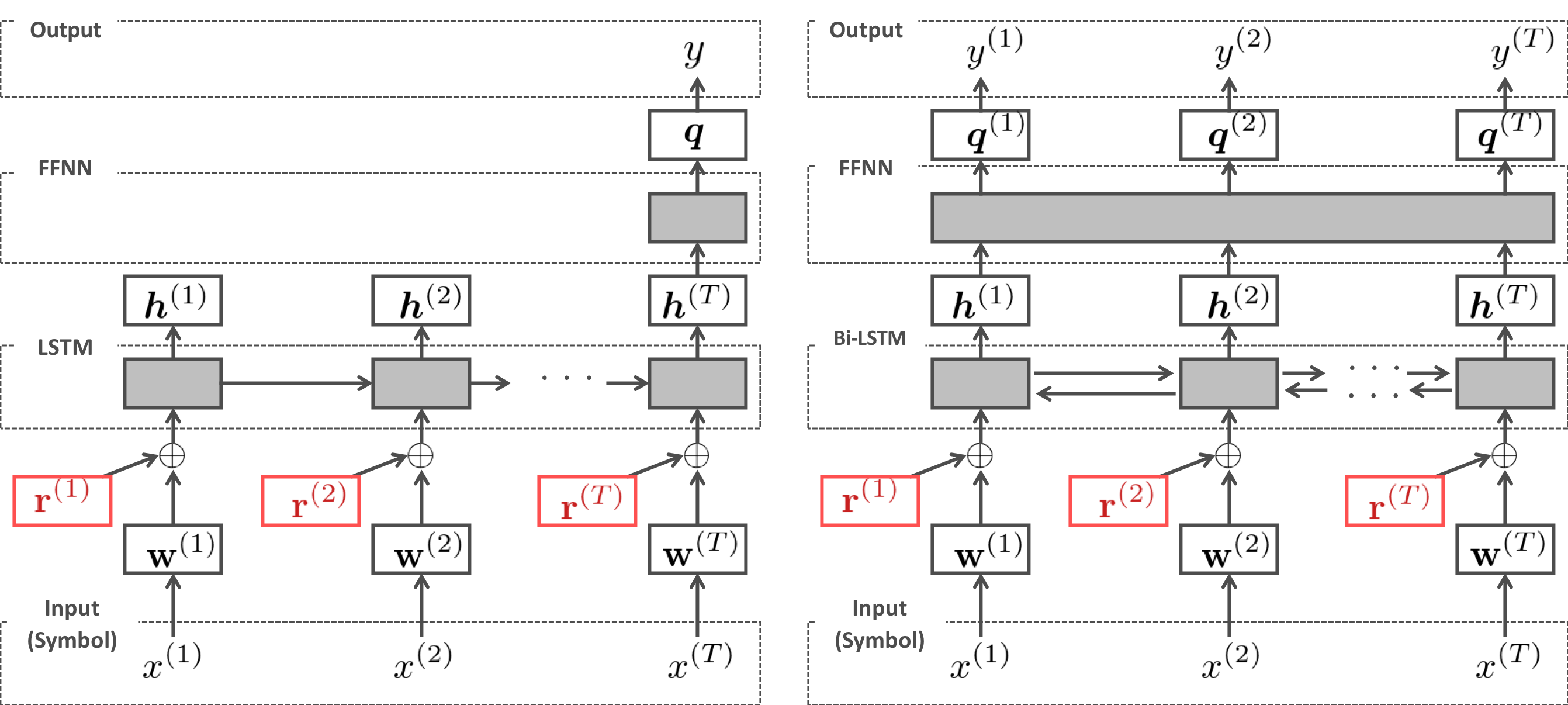}
\caption{Overview of our neural models with perturbation: $\bm{r}$ denotes the perturbation, which is $\bm{r}^{(t)}_{\tt AdvT}$, $\bm{r}^{(t)}_{\tt VAT}$, $\bm{r}(\bm{{\alpha}_{\tt iAdvT}}^{(t)})$, or $\bm{r}(\bm{{\alpha}_{\tt VAT}}^{(t)})$, depending on the method.}
\label{fig:lstm_baseline_overview_with_perturbation}
\end{center}
\end{figure}

%%%%%%%%%%%%%%%%%%%%%%%%%%%%%
{\it Adversarial training} (AdvT)~\cite{Goodfellow2014ExplainingAH} is a novel regularization method that improves the robustness of misclassifying small perturbed inputs.
To distinguish the AdvT method in image processing, this paper specifically refers to AdvT that is applied to input word embedding space for NLP tasks as AdvT-Text, which
was first introduced in~\cite{miyato2016adversarial}.
Let $\bm{r}^{(t)}_{\tt AdvT}$ be a (adversarial) perturbation vector for $t$-th word $x^{(t)}$ in input $\tilde{X}$.
We assume that $\bm{r}^{(t)}_{\tt AdvT}$ is a $D$-dimensional vector whose dimension always matches that of word embedding vector $\bm{w}^{(t)}$.
Fig.~\ref{fig:lstm_baseline_overview_with_perturbation} shows the AdvT-Text architecture and our baseline neural models by applying AdvT.
See also Fig.~\ref{fig:lstm_baseline_overview} for a comparison of the architecture with our baseline neural models.
%
%%%%%%%%%%%%%%%%%%%%%%%%%%%%%%%%%%%%%%%%%%%%%%%%%%%%%%
%%%%%%%%%%%%%%%%%%%%%%%%%%%%%%%%%%%%%%%%%%%%%%%%%%%%%%
%
Let $\bm{r}$ represent a concatenated vector of $\bm{r}^{(t)}$ for all $t$.
We introduce ${\tilde{X}}_{+\bm{r}}$ that denotes $\tilde{X}$ with additional small perturbations, where
$ {\tilde{X}}_{+\bm{r}} = (\bm{w}^{(t)}+ \bm{r}^{(t)})^{T}_{t=1}$.
To obtain (worst case) perturbations $\bm{r}^{(t)}_{\tt AdvT}$ for all $t$ for maximizing the negative log-likelihood (equivalent to minimizing the log-likelihood), we seek optimal solution $\bm{r}_{\tt AdvT}$ by maximizing the following equation:
\begin{align}
 \bm{r}_{\tt AdvT}
 =&
 \argmax_{\bm{r}, ||\bm{r}||\le \epsilon} \Big\{ \ell ( {\tilde{X}}_{+\bm{r}}  , \tilde{Y}, \mathcal{W}) \Big\}
 ,
\label{eq:adv_maximize}
\end{align}
where $\epsilon$ is a tunable hyper-parameter that controls the norm of the perturbation
and $\bm{r}_{\tt AdvT}$ represents a concatenated vector of $\bm{r}^{(t)}_{\tt AdvT}$ for all $t$ that resemble $\bm{r}$.
Then based on adversarial perturbation $\bm{r}_{\tt AdvT}$, the loss function for AdvT-Text can be defined:
\begin{align}
 \mathcal{J}_{{\tt AdvT}} (\mathcal{D},\mathcal{W})
 &=
 \frac{1}{|\mathcal{D}|} \sum_{(\tilde{X},\tilde{Y})\in \mathcal{D}} \ell( \tilde{X}_{+\bm{r}_{\tt AdvT}},\tilde{Y}, \mathcal{W})
 ,
 \label{eq:adv_loss}
\end{align}
where ${\tilde{X}}_{+\bm{r}_{\tt AdvT}} = (\bm{w}^{(t)}+ \bm{r}^{(t)}_{\tt AdvT})^{T}_{t=1}$, similar to $\tilde{X}_{+\bm{r}}$.
%

%%%%%%%%%%%%%%%%%%%
It is generally infeasible to exactly estimate $\bm{r}_{\tt AdvT}$ in Eq.~\ref{eq:adv_maximize} for sophisticated deep neural models.
 As a solution, an approximation method was proposed by linearizing $\ell ({\tilde{X}}, \tilde{Y}, \mathcal{W})$ around ${\tilde{X}}$~\cite{Goodfellow2014ExplainingAH}.
 For our RNN-based models, the approximation method induces the following non-iterative solution for calculating $\bm{r}^{(t)}_{\tt AdvT}$ for all $t$:
 \begin{align}
  \bm{r}^{(t)}_{\tt AdvT}
  =&
  \frac{\epsilon \bm{g}^{(t)}} {||\bm{g}||_2}
  ,\quad
  \bm{g}^{(t)} = \nabla_{\bm{w}^{(t)}} \ell (\tilde{X},\tilde{Y}, \mathcal{W})
  ,
 \label{eq:adv_approx}
 \end{align}
where $\bm{g}$ is a concatenated vector of $\bm{g}^{(t)}$ for all $t$.

%%%%%%%%%%%%%%%%%%%
Finally, we jointly minimize objective functions $\mathcal{J} (\mathcal{D},\mathcal{W})$ and $\mathcal{J}_{{\tt AdvT}} (\mathcal{D},\mathcal{W})$:
\begin{align}
 \hat{\mathcal{W}} &=  \argmin_{\mathcal{W}} \Big\{ \mathcal{J} (\mathcal{D}, \mathcal{W}) + \lambda \mathcal{J}_{{\tt AdvT}} (\mathcal{D}, \mathcal{W}) \Big\}
 ,
\label{eq:obj_advt}
\end{align}
where $\lambda$ is a coefficient that controls the balance of two loss functions.

%%%%%%%%%%%%%%%%%%%%%%%%%%%%%%%%%%%%%%%%%%%%%%%%%%%%%%%%
%%%%%%%%%%%%%%%%%%%%%%%%%%%%%%%%%%%%%%%%%%%%%%%%%%%%%%%%
%%%%%%%%%%%%%%%%%%%%%%%%%%%%%%%%%%%%%%%%%%%%%%%%%%%%%%%%
\section{Interpretable Adversarial Perturbation}

As described above, we extended Adv-Text to restore the ability to generate adversarial texts that are interpretable by people while maintaining the task performance.
We only restrict the directions of the perturbations in the embedding space toward existing words in the input word embedding space.
The intuition behind our method is that the directions to other words can be interpreted as the substitution of another word in the sentence, which may reconstruct adversarial texts.
We refer to our AdvT-Text extension as {\it interpretable AdvT-Text} or iAdvT-Text.

\subsection{Definition of interpretable AdvT-Text}

Suppose $\bm{w}_{k}$ denotes a word embedding vector that corresponds to the $k$-th word in vocabulary $\mathcal{V}$.
We define {\it direction vector} ${\bm{d}}^{(t)}_{k}$ that indicates the direction from $\bm{w}^{(t)}$ to $\bm{w}_{k}$ in the input word embedding space:
\begin{align}
 {\bm{d}}^{(t)}_{k} = \frac{\tilde{\bm{d}}^{(t)}_{k}}{||\tilde{\bm{d}}^{(t)}_{k}||_2 }
 , \quad \mbox{where}\quad
 \tilde{\bm{d}}^{(t)}_{k} = \bm{w}_{k} - \bm{w}^{(t)}
 \label{eq:adv_direction}
 .
\end{align}
Note that ${\bm{d}}^{(t)}_{k}$ for all $t$ and $k$ is always a unit vector, $||{\bm{d}}^{(t)}_{k}||_2=1$.
If the $t$-th word in the given input sentence is the $k$-th word in the vocabulary, then $\bm{w}_{k} = \bm{w}^{(t)}$, and thus, $\bm{d}^{(t)}_{k}$ becomes a zero vector\footnote{If $\tilde{\bm{d}}^{(t)}_{k} = \bm{0}$, then we treat ${\bm{d}}^{(t)}_{k} = \bm{0}$.}.

Next let $\bm{\alpha}^{(t)}$ be a $|\mathcal{V}|$-dimensional vector, and let $\alpha^{(t)}_k$ be the $k$-th factor of $\bm{\alpha}^{(t)}$, where $\bm{\alpha}^{(t)} = (\alpha^{(t)}_k)^{|\mathcal{V}|}_{k=1}$.
We define $\bm{r}(\bm{\alpha}^{(t)})$ that denotes the perturbation generated for the $t$-th word in $\tilde{X}$, which is parameterized by $\bm{\alpha}^{(t)}$:
\begin{align}
 \bm{r}(\bm{\alpha}^{(t)}) =& {\sum}^{|\mathcal{V}|}_{k=1} \alpha^{(t)}_k \bm{d}^{(t)}_{k}
 .
 \label{eq:advR}
\end{align}
$\alpha^{(t)}_k$ is a weight for the direction from the $t$-th word in the input to the $k$-th word in the vocabulary.
Then, similar to the definition of $ {\tilde{X}}_{+\bm{r}}$,
we also introduce $ {\tilde{X}}_{+\bm{r}(\bm{\alpha})}$ as follows:
\begin{align}
% {\tilde{X}}_{+\bm{r}_{\tt iAdvT}} = \big( \bm{w}^{(t)} + \bm{r}^{(t)}_{\tt iAdvT}\big)^{T}_{t=1}
 {\tilde{X}}_{+\bm{r}(\bm{\alpha})}
 =
 \big( \bm{w}^{(t)} + \bm{r}(\bm{\alpha}^{(t)}) \big)^{T}_{t=1}
 .
 \label{eq:advR_seq}
\end{align}
%

%%%%%%%%%%%%%%%%%%%%%%%%%
Similar to Eq.~\ref{eq:adv_maximize},
 we seek the worst case weights of the direction vectors that maximize the loss functions
 as follows:
 \begin{align}
 \bm{\alpha}_{\tt iAdvT}
 =&
 \argmax_{\bm{\alpha}, ||\bm{\alpha}||\le \epsilon} \Big\{ \ell (  {\tilde{X}}_{+\bm{r}(\bm{\alpha})}, \tilde{Y}, \mathcal{W}) \Big\}
  \label{eq:advR_maximize}
 .
\end{align}
Then we define the loss functions of our method, iAdvT-Text, based on $\bm{\alpha}_{\tt iAdvT}$:
\begin{align}
 \mathcal{J}_{{\tt iAdvT}} (\mathcal{D},\mathcal{W})
 &=
 \frac{1}{|\mathcal{D}|} \sum_{(\tilde{X},\tilde{Y})\in \mathcal{D}} \ell( \tilde{X}_{+\bm{r}(\bm{\alpha}_{\tt iAdvT})},\tilde{Y}, \mathcal{W})
 \label{eq:advR_loss}
.
\end{align}
%
%%%%%%%%%%%%%%%%%%%
We substitute $\mathcal{J}_{{\tt AdvT}} (\mathcal{D},\mathcal{W})$ in Eq.~\ref{eq:obj_advt} with $\mathcal{J}_{{\tt iAdvT}} (\mathcal{D},\mathcal{W})$ for our method, where the form of the optimization problem can be simply written:
\begin{align}
 \hat{\mathcal{W}} &=  \argmin_{\mathcal{W}} \Big\{ \mathcal{J} (\mathcal{D}, \mathcal{W}) + \lambda \mathcal{J}_{{\tt iAdvT}} (\mathcal{D}, \mathcal{W}) \Big\}
 .
\end{align}

To reduce the calculation cost, we also introduce an update formula derived by applying the same idea of the approximation method explained in Eq.~\ref{eq:adv_approx}:
\begin{align}
 \bm{\alpha}^{(t)}_{\tt iAdvT}
 =
 \frac{\epsilon\bm{g}^{(t)}}{||\bm{g}||_2}
 ,
 \,\,\,
 \bm{g}^{(t)}
 =&
 \nabla_{\bm{\alpha}^{(t)}} \ell({\tilde{X}}_{+\bm{r}(\bm{\alpha})},\tilde{Y}, \mathcal{W})
.
 \label{eq:update_alpha_new}
\end{align}
%
%%%%%%%%%%%%%%%%%%%
Similar to $\bm{r}^{(t)}_{\tt AdvT}$,
the intuitive interpretation of $\bm{\alpha}^{(t)}_{\tt iAdvT}$ is the (normalized) strength of each direction $\bm{d}^{(t)}_k$ about how much to increase the loss function.
 Thus, we expect to evaluate which direction of words is a good adversarial perturbation.
%%%%%%%%%%%%%%%%%

\subsection{Practical computation}

The most time-consuming part of our method is its calculation of the summation of all the words that appeared in Eq.~\ref{eq:advR}, which includes the calculation of directions $\bm{d}^{(t)}_{k}$ for all the words from each word $\bm{w}^{(t)}$, as shown in Eqs.~\ref{eq:adv_direction}.
At most, this creates a computational cost of $|\mathcal{V}|^2$, which might be unacceptable compared with the small computational cost of AdvT-Text (the previous method).
Here we introduce $\mathcal{V}^{(t)}$ as individual vocabularies of step $t$, where $ \mathcal{V}^{(t)} \subseteq \mathcal{V}$ for all $t$ and $ |\mathcal{V}^{(t)} |\ll |\mathcal{V}|$, i.e., $ |\mathcal{V}^{(t)}|=10$.
In our method, we select the $|\mathcal{V}^{(t)}|$ nearest neighbor word embeddings around each $\bm{w}^{(t)}$ for all $t$ in each iteration during the training.
This approximation is equivalent to treating $\alpha^{(t)}_k=0$ for all $k$ if $w_k \notin \mathcal{V}^{(t)}$ for all $t$.
 The intuition behind this approximation is that words with a large distance can be treated as nearly unrelated words.

\subsection{Extension to semi-supervised learning}
Suppose $\mathcal{D}'$ denotes a set of labeled and unlabeled data.
{\it Virtual adversarial training} (VAT)~\cite{Miyato2015DistributionalSW} is a (regularization) method closely related to AdvT.
VAT, a natural extension of AdvT to semi-supervised learning, can also be applied to tasks in NLP fields, which we refer to as VAT-Text.
We borrow this idea and extend it to our iAdvT-Text for a semi-supervised setting, which we refer to as iVAT-Text.

%%%%%%%%%%%%%%%%%
VAT-Text uses the following objective function for estimating the loss of adversarial perturbation $\bm{r}_{\tt VAT}$:
\begin{align}
 \mathcal{J}_{{\tt VAT}} (\mathcal{D}',\mathcal{W}) &= \frac{1}{|\mathcal{D}'|} \sum_{\tilde{X}\in \mathcal{D}'}  \ell_{{\tt KL}} (\tilde{X}, \tilde{X}_{+\bm{r}_{\tt VAT}} \mathcal{W})
 \label{eq:vat_loss}
 \\
 \ell_{{\tt KL}} (\tilde{X}, \tilde{X}_{+\bm{r}_{\tt VAT}}, \mathcal{W})
 &=
 {\tt KL} \big( p( \cdot \mid {\tilde{X}}, {\mathcal{W}}) || p( \cdot \mid \tilde{X}_{+\bm{r}_{\tt VAT}}, \mathcal{W}) \big)
 \nonumber
 ,
\end{align}
where ${\tt KL}(\cdot||\cdot)$ denotes the KL divergence.
To obtain $\bm{r}_{\tt VAT}$, we solve the following optimization problem:
\begin{align}
 \bm{r}_{\tt VAT}
 &=
 \argmax_{\bm{r}, ||\bm{r}||\le \epsilon}
 \Bigr\{
 {\tt KL} \big( p( \cdot \mid {\tilde{X}}, {\mathcal{W}}) || p( \cdot \mid \tilde{X}_{+\bm{r}}, {\mathcal{W}}) \big)
 \Bigr\}
 .
 \label{eq:vat_obj}
\end{align}
Then instead of solving the above optimization problem, an approximated method was proposed~\cite{miyato2016adversarial}:
 \begin{align}
  \bm{r}^{(t)}_{\tt VAT} =& \frac{\epsilon \bm{g}^{(t)}}{||\bm{g}||_2}
  ,\quad
  \bm{g}^{(t)} = \nabla_{\bm{w}^{(t)}+\bm{r}^{(t)}} \ell_{\tt KL} (\tilde{X},\tilde{X}_{+\bm{r}},\mathcal{W})
  .
 \label{eq:vat_approx}
\end{align}

By using the same derivation technique to obtain the above approximation, we introduce the following equation to calculate $\bm{\alpha}^{(t)}_{\tt iVAT}$ for an extension to semi-supervised learning:
\begin{align}
 \bm{\alpha}^{(t)}_{\tt iVAT}
 =
 \frac{\epsilon\bm{g}^{(t)}}{||\bm{g}||_2}
 ,
 \quad
 \bm{g}^{(t)}
 =&
 \nabla_{\bm{\alpha}^{(t)}} \ell_{\tt KL}({\tilde{X}}, {\tilde{X}}_{+\bm{r}(\bm{\alpha})}, \mathcal{W})
.
 \label{eq:update_alpha_new2}
\end{align}
Then the objective function for iVAT-Text can be written:
\begin{align}
 \mathcal{J}_{{\tt iVAT}} (\mathcal{D}',\mathcal{W}) &= \frac{1}{|\mathcal{D}'|} \sum_{\tilde{X}\in \mathcal{D}'}  \ell_{{\tt KL}} (\tilde{X}, \tilde{X}_{+\bm{r}(\bm{\alpha}_{\tt iVAT})} \mathcal{W})
 \label{eq:ivat_loss}
 .
\end{align}

%%%%%%%%%%%%%%%%%%%%%%%%%%%%%%%%%%%%
%%%%%%%%%%%%%%%%%%%%%%%%%%%%%%%%%%%%
%%%%%%%%%%%%%%%%%%%%%%%%%%%%%%%%%%%%
\begin{table}[t]
 \centering
 \caption{\label{tab:dataset_statics} Summary of datasets}
 \tabcolsep 3pt
 {\small
   \begin{tabular}{llrrrr}
    \hline
       Task & Dataset  & Train & Dev & Test & Unlabeled \\ \hline
       \multirow{3}{*}{SEC} & IMDB         & 21,246  &  3,754 &   25,000  &  50,000  \\
                            & Elec         & 22,500  &  2,500 &   25,000  & 200,000  \\
                            & Rotten Tomatoes  & 8,636   &  960  &  1,066   & 7,911,684   \\\hline
       \multirow{2}{*}{CAC} & DBpedia      & 504,000  &  56,000  &  70,000    &   -      \\
                            & RCV1         & 14,007  &  1,557 &    49,838 &  668,640      \\ \hline
       GED  & FCE-public   & 28,731  &  2,222 &    2,720  &   -    \\
   \hline
   \end{tabular}
 }
\end{table}
%%%%%%%%%%%%%%%%%%%%%%%%%%%%%%%%%%%%
\begin{table}[t]
 \centering
 \caption{\label{tab:hyperparam} Summary of hyper-parameters}
 \tabcolsep 3pt
 {\small
    \begin{tabular}{l|l|c|c|c}
     \hline
     \  & Hyper-parameter          & SEC & CAC & GED  \\ \hline

       \multirow{2}{*}{Word embed.}
              & dimensions         & \multicolumn{2}{c|}{256} & 300 \\
              & dropout            & \multicolumn{3}{c}{0.5}  \\ \hline
       \multirow{2}{*}{LSTM}
              & state size               & \multicolumn{2}{c|}{1024}  & 200  \\
              & direction                & \multicolumn{2}{c|}{Uni-LSTM} & Bi-LSTM \\ \hline
       \multirow{2}{*}{FFNN}
             & dimensions               & 30 & 128 & 50  \\
             & activation               & \multicolumn{3}{c}{ReLU}   \\ \hline
      \multirow{4}{*}{Optimization}
              & algorithm                & \multicolumn{3}{c}{Adam} \\
              & batch size               & \multicolumn{3}{c}{32} \\
              & initial learning rate    & \multicolumn{3}{c}{0.001} \\
              & decay rate               & \multicolumn{3}{c}{0.9998} \\
    \hline
   \end{tabular}
 }
\end{table}
%%%%%%%%%%%%%%%%%%%%%%%%%%%%%%%%%%%%

\section{Experiments}
We conducted our experiments on a sentiment classification (SEC) task, a category classification (CAC) task, and a grammatical error detection (GED) task to evaluate the effectiveness of our methods, iAdvT-Text and iVAT-Text.
SEC is a text classification task that classifies a given text into either a positive or a negative class.
GED is a sequence labeling task that identifies ungrammatical words.

\subsection{Datasets}
For SEC, we used the following well-studied benchmark datasets, IMDB~\cite{Maas2011LearningWV}, Elec~\cite{Johnson2015SemisupervisedCN}, and Rotten Tomatoes~\cite{Pang2005SeeingSE}.
In our experiment with the Rotten Tomatoes dataset, we utilized unlabeled examples from the Amazon Reviews dataset\footnote{\url{http://snap.stanford.edu/data/web-Amazon.html}}.
For CAC, we utilized DBpedia~\cite{Lehmann2015DBpediaA} and RCV1~\cite{Lewis2004RCV1AN}.
Since the DBpedia dataset has no additional unlabeled examples, the DBpedia results are only for the supervised learning task.
Following ~\protect\cite{miyato2016adversarial}, we split the original training data into training and development sentences.
For GED, we utilized the First Certificate in the English dataset (FCE-public)~\cite{yannakoudakis2011new}.
Table \ref{tab:dataset_statics} summarizes the information about each dataset.

%%%%%%%%%%%%%%%%%%%%%%%%%%%%%%%%%%%%
\begin{table}[t]
 \centering
 \caption{\label{tab:result} Test performance (error rate) on IMDB: lower is better. Semi-supervised learning models are marked with \ding{61}.}
  \tabcolsep 3pt
 {\small
   \begin{tabular}{l|r}
    \hline
    Method             & Test error rate \\
    \hline
       Baseline                       &  7.05  (\%) \\
       Random Perturbation (Labeled) &  6.74  (\%) \\
       iAdvT-Rand (Ours)  &  6.69  (\%) \\
       iAdvT-Best (Ours)  &  6.64  (\%) \\
       AdvT-Text~\cite{miyato2016adversarial}  &  6.12  (\%) \\
       \bf iAdvT-Text (Ours)  &  \bf 6.08  (\%) \\
    \hline
       Random Perturbation (Labeled + Unlabeled)\ding{61} &  6.44  (\%) \\
       iVAT-Rand (Ours)\ding{61}  &  6.08  (\%) \\
       iVAT-Best (Ours)\ding{61}  &  6.30  (\%) \\
       VAT-Text~\cite{miyato2016adversarial}\ding{61}    &  5.69  (\%) \\
       \bf iVAT-Text (Ours)\ding{61}  &  \bf 5.66  (\%) \\
    \hline
       Full+Unlabeled+BoW \cite{Maas2011LearningWV}  & 11.11  (\%)   \\
       Paragraph Vectors \cite{Le2014DistributedRO}   &  7.42  (\%)   \\
       SA-LSTM \cite{Dai2015SemisupervisedSL}\ding{61}         &  7.24  (\%)   \\
       One-hot bi-LSTM \cite{Johnson2016SupervisedAS}\ding{61}     &  5.94  (\%)   \\
    \hline
  \end{tabular}
  }
\end{table}
%%%%%%%%%%%%%%%%%%%%%%%%%%%%%%%%%%%%
\begin{table}[t]
 \centering
 \caption{\label{tab:result_elec} Test performance (error rate) on Elec, RCV1, and Rotten Tomatoes: lower is better. Semi-supervised learning models are marked with \ding{61}.}
  \tabcolsep 3pt
 {\small
 %\begin{center}
   \begin{tabular}{l|c|c|c} %\toprule
    \hline
       Method                                           & Elec           & RCV1      & Rotten\\ \hline
       Baseline                                         & 6.24 (\%)      & 12.01 (\%) & 17.36  (\%) \\
       AdvT-Text~\cite{miyato2016adversarial}           & 5.94 (\%)      & 10.93 (\%) & 15.84  (\%) \\
       \bf iAdvT-Text (Ours)                            &  \bf 5.58 (\%) & \bf 10.07 (\%) & \bf 14.24  (\%)  \\ \hline
       VAT-Text~\cite{miyato2016adversarial}\ding{61}   &  5.66    (\%)  & 11.80 (\%) & 14.26  (\%) \\
       \bf iVAT-Text (Ours)\ding{61}                    &  \bf 5.18 (\%) & \bf 11.68 (\%) & \bf 14.12 (\%)\\
       % Random (Labeled)                               &    (\%) &    (\%) &    (\%)  \\
       % Random (Labeled + Unlabeled)\ding{61}          &     (\%) &    (\%)  &    (\%)  \\
    \hline%\toprule
  \end{tabular}
  }
\end{table}
%%%%%%%%%%%%%%%%%%%%%%%%%%%%%%%%%%%%
\begin{table}[t]
 \centering
 \caption{\label{tab:result_dbpedia} Test performance (error rate) on DBpedia: lower is better}
  \tabcolsep 3pt
 {\small
   \begin{tabular}{l|r}
    \hline
       Method             & Test error rate \\ \hline
       Baseline                                     & 0.94   (\%) \\
       % Random Perturbation (Labeled)  &    (\%) \\
       AdvT-Text~\cite{miyato2016adversarial}   & \bf 0.92    (\%) \\
       \bf iAdvT-Text (Ours)                            &  0.99 (\%) \\ \hline
       VAT-Text~\cite{miyato2016adversarial}    &  \bf  0.91 (\%) \\
       \bf iVAT-Text (Ours)  &  0.93   (\%) \\
    \hline%\toprule
  \end{tabular}
  }
\end{table}
%%%%%%%%%%%%%%%%%%%%%%%%%%%%%%%%%%%%
\begin{table}[t]
 \centering
 \caption{\label{tab:result_grammar} Test performance ($F_{0.5}$) on GED task: larger is better}
 \tabcolsep 3pt
 {\small
   \begin{tabular}{l|r}
    \hline
       Method             &$F_{0.5}$\\ \hline
       Baseline                                    & 39.21    \\
       Random Perturbation                         & 39.90    \\
       AdvT-Text~\cite{miyato2016adversarial}      & \bf 42.28   \\
       \bf iAdvT-Text (Ours)             & 42.26   \\
       VAT-Text~\cite{miyato2016adversarial} & 41.81 \\
       \bf iVAT-Text (Ours)     & 41.88 \\ \hline
       BiLSTM w/{\tt Skipgram}~\cite{rei-yannakoudakis:2016:P16-1} & 41.1\,\,\,\\
       BiLSTM w/{\tt GWE}~\cite{kaneko2017grammatical}              & 41.4\,\,\,\\
   \hline
  \end{tabular}
 }
\end{table}

\subsection{Model settings}

To fairly compare our methods with previous methods,
we followed previously described model configurations~\cite{miyato2016adversarial} for SEC and \cite{rei-yannakoudakis:2016:P16-1,kaneko2017grammatical} GED,
shown in Fig. \ref{fig:lstm_baseline_overview_with_perturbation}: left for SEC and right for GED.
Moreover, following~\cite{miyato2016adversarial}, we initialized the word embeddings and the LSTM weights with a pre-trained RNN-based language model~\cite{Bengio2000ANP} that was trained on labeled training and unlabeled data if they were available.
To reduce the computational cost of softmax loss, we use the Adaptive Softmax \cite{Grave2017EfficientSA} for training language model.
We utilized an early stopping criterion~\cite{Caruana2000OverfittingIN} based on the performance measured on development sets.
The hyper-parameters are summarized in Table \ref{tab:hyperparam}, with dropout~\cite{Srivastava2014DropoutAS} and Adam~\cite{Kingma2014AdamAM}.
In addition, we set $\epsilon=5.0$ for both AdvT-Text and VAT-Text and $\epsilon=15.0$ for our method.
We also set $\lambda=1$ for all the methods.
To find the best hyper-parameter, we picked models whose performances were best measured on development data.

In addition, we implemented our methods (iAdvT-Text and iVAT-Text) and re-implemented the previous methods (AdvT-Text and VAT-Text) using Chainer \cite{chainer_learningsys2015} with GPU support.
All four methods share sub-modules, such as RNN-based models, in our implementation.
Therefore, our internal experiments are fairly compared under identical conditions.

%%%%%%%%%%%%%%%%%%%%%%%%%%%%%%%%%%
\begin{figure}[t]
\begin{center}
\includegraphics[width=1.0\linewidth]{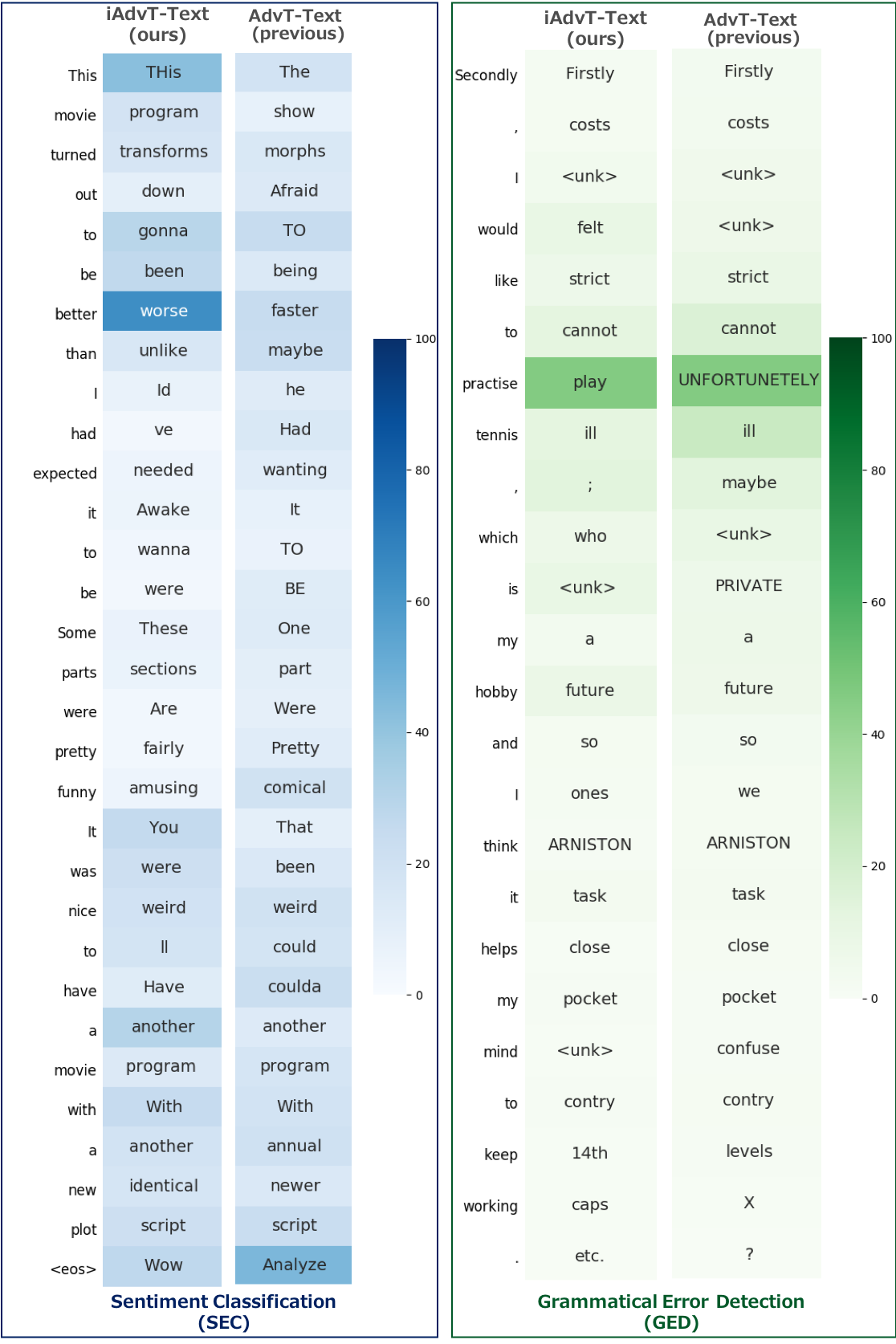}
\caption{Visualization of perturbation at sentence-level: Texts at left of blue or green bars are sentences in datasets, and texts in blue or green bars are words reconstructed from perturbations.}
\label{fig:visualizing_example_at_sentence_level}
\end{center}
\end{figure}
%%%%%%%%%%%%%%%%%%%%%%%%%%%%%%%%%%

%%%%%%%%%%%%%%%%%%%%%%%%%%%%%%%%%%%%%%%%%%%%%%%%%%%%%
%%%%%%%%%%%%%%%%%%%%%%%%%%%%%%%%%%%%%%%%%%%%%%%%%%%%%
%%%%%%%%%%%%%%%%%%%%%%%%%%%%%%%%%%%%%%%%%%%%%%%%%%%%%
\subsection{Evaluation by task performance}

\begin{table*}[t]
 \centering
 \caption{Adversarial examples, misclassified by trained model and reconstructed by iAdvT-Text}
 \label{tab:result_adv_example}
 \tabcolsep 2pt
 {\scriptsize
   \begin{tabular}{p{1.2cm}|p{14.5cm}|c}
    \hline
                    & Sentence  (SEC)               &  Prediction  \\ \hline\hline
      Original Sentence
        & {The essence \colorbox{yellow}{\underline{of}} \colorbox{yellow}{\underline{this}} film falls on judgments by police officers who fortunately ethical and moral men act on situations within situations in a city with a super abundance of violence and killing Good compound interacting story lines and above average characterizations \verb|<|eos\verb|>| }
        &  \colorbox{pink}{Positive} \\  \hline
      Adversarial Example
        & {The essence \colorbox{yellow}{\underline{from}} \colorbox{yellow}{\underline{THIS}} film falls on judgments by police officers who fortunately ethical and moral men act on situations within situations in a city with a super abundance of violence and killing Good compound interacting story lines and above average characterizations \verb|<|eos\verb|>| }
         &  \colorbox{cyan}{Negative} \\ \hline \hline

      Original Sentence
          & {There is really but one thing to say about \colorbox{yellow}{\underline{this}} sorry movie It should never have been made The first one one of my favourites An American Werewolf in London is a great movie with a good plot good actors and good FX But this one It stinks to heaven with a cry of helplessness \verb|<|eos\verb|>|  }
          &  \colorbox{cyan}{Negative} \\  \hline
      Adversarial Example
          & {There is really but one thing to say about \colorbox{yellow}{\underline{that}} sorry movie It should never have been made The first one one of my favourites An American Werewolf in London is a great movie with a good plot good actors and good FX But this one It stinks to heaven with a cry of helplessness \verb|<|eos\verb|>|  }
          &  \colorbox{pink}{Positive} \\ \hline

   \end{tabular}
   \vskip 1.em
   \begin{tabular}{p{2.5cm}|p{9.0cm}|l|c}
     \hline
                     & Sentence  (GED)            &  Prediction   & Correct Replacement \\ \hline\hline
        Original Sentence
          & {We all want to thank you for having \colorbox{pink}{\underline{choose}} such good places in London .}
          & {\scriptsize 0 0 0 0 0 0 0 0 \colorbox{pink}{1} 0 0 0 0 0 0}
          &   \\  \hline
        Adversarial Example
          & {We all want to thank you for having \colorbox{yellow}{\underline{choosing}} such good places in London .}
          &  {\scriptsize 0 0 0 0 0 0 0 0 \colorbox{yellow}{0} 0 0 0 0 0 0}
          & (choose $\to$  chosen) \\ \hline \hline
       Original Sentence
           & {I am not really satisfied \colorbox{pink}{\underline{about}} it .}
           &  {\scriptsize 0 0 0 0 0 \colorbox{pink}{1} 0 0}
           & \\  \hline
       Adversarial Example
           & {I am not really satisfied \colorbox{yellow}{\underline{more}} it .}
           &  {\scriptsize 0 0 0 0 0 \colorbox{yellow}{0} 0 0 }
           &  (about $\to$ with)
            \\ \hline
   \end{tabular}
  }
\end{table*}
%%%%%%%%%%%%%%%%%%%%%%%%%%%%%%%%%%%

Table \ref{tab:result} shows the IMDB performance evaluated by the error rate.
Random perturbation (Labeled) is the method with which we replaced $\bm{r}^{(t)}_{\tt AdvT}$ with a random unit vector, and Random Perturbation (Labeled + Unlabeled) is the method with which we replaced $\bm{r}^{(t)}_{\tt VAT}$ with a random unit vector.
We tried two simple methods, iAdvT-Rand and iAdvT-Best.
iAdv-Rand is the method with which we replaced $\bm{r}^{(t)}_{\tt AdvT}$ with the nearest ten, randomly picked words vectors.
iAdv-Best is the method from which we picked the best direction based on Eq.~\ref{eq:advR_seq}.
%%%%%%%
Surprisingly, iAdvT-Text outperformed AdvT-Text,
and iVAT-Text achieved the same performance level and slightly outperformed VAT-Text%
\footnote{The AdvT-Text and VAT-Text scores were obtained by our re-implemented code, which outperformed the original scores \cite{miyato2016adversarial} (Adv:6.21 \%, VAT:5.91 \%).}.
Note here that our method was mainly designed to restore the interpretability, not to improve the task performance.
Before evaluating our methods, iAdvT-Text and iVAT-Text, we assumed that they would respectively degrade the AdvT-Text and VAT-Text performances, since our methods strongly restrict the degrees of freedom for the direction of the perturbations for interpretability.
This suggests that the directions of the existence of actual words in the word embedding space provide useful information for improving the generalization performance.
The performance of two simple methods (iAdvT-Rand and iAdvT-Best) is poor.
Tables \ref{tab:result_elec} and \ref{tab:result_dbpedia} show the performance on the other datasets\footnote{For RCV1 and DBpedia, our baselines were slightly weak due to the resource limitation of constracting the large-scale pre-trainined language models.}.

%%%%%%%%%%%%%
Table \ref{tab:result_grammar} shows the test performance on the GED task.
We used $F_{0.5}$ as an evaluation measure for GED,
which was adopted in the CoNLL-14 shared task \cite{Rozovskaya2014TheIS}%
\footnote{
A previous study~\cite{Nagata2010EvaluatingPO} suggested that since accurate prediction is more important than coverage in error detection applications, $F_{0.5}$ was selected rather than $F_{1}$.}.
Other reported results~\cite{rei-yannakoudakis:2016:P16-1,kaneko2017grammatical} are around the current state-of-the-art performance on this dataset.
In our experiments,
AdvT-Text achieved the highest $F_{0.5}$.
Our experiments revealed that AdvT-Text can further outperform the current state-of-the-art methods.
Moreover, our methods successfully reached performances that almost matched AdvT-Text and VAT-Text.
Again, we emphasize that these results are substantially positive for our methods since they did not degrade the performance even when we added a strong restriction for calculating the perturbations.

In addition, in contrast to SEC, VAT-Text did not outperform AdvT-Text.
Since the GED dataset does not contain a large amount of unlabeled data, we confirmed that it is hard for VAT-Text to improve the performance.

\subsection{Visualization of sentence-level perturbations}
We visualized the perturbations computed by our method (iAdvT-Text) in Fig.~\ref{fig:visualizing_example_at_sentence_level} for understanding its behavior.
We also visualized the perturbations by the previous method (AdvT-Text) for comparison.
The words at the left of each (blue or green) bar indicate the words in the (true) sentences in the dataset.
We selected the highest direction toward a word from each word in the sentence.
In our method, it can be easily obtained by selecting the maximum values of $\alpha^{(t)}_k$ for all $t$.
For AdvT-Text, we calculated the cosine similarities between the perturbation and direction to every word $\bm{w}_k$ and selected a word with the highest cosine similarity.
Each word written in the (blue or green) bar represents the selected word by the above operations, and shades of color are the relative strengths of the obtained perturbations toward the selected words.

For the SEC task, the correct label of the sentence in the blue bar is {\it positive}.
The iAdvT-Text successfully found the directions for replacing {\it better} with {\it worse} to increase the loss.
In other words, the direction might change the class label from positive to negative.
For the GED task, the sentence in the green bar contains a grammatical error word ({\it practise}), which should be replaced with {\it play}.
The iAdvT-Text also found directions for replacing {\it practise} with {\it play}.

In contrast, the perturbations of AdvT-Text (Previous) were uninterpretable (replacing {\it \verb|<|eos\verb|>|} with {\it Analyze}, and replacing {\it practise} with {\it UNFORTUNETELY}).
This is mainly because the perturbations of AdvT-Text barely matched the direction toward any existence points of word embeddings, and we just visualized the most cosine similar words with perturbation.

These results revealed that the directions of the perturbations in iAdvT-Text are understandable by humans, and thus, offer a chance for researchers to interpret black-box neural models, regardless whether the model properly learned certain phenomena that the researchers are interested in.
We believe that such interpretability is critical, especially for sophisticated neural models.
The usefulness of this visualization is the main claim of our proposed methods.

\subsection{Adversarial texts}
We reconstructed adversarial examples, which misclassified the trained models, from the adversarial perturbations in the input word embedding space given by iAdvT-Text.
To obtain adversarial texts, we first identified the largest perturbation and replaced the original word with one that matches the largest perturbation.

Table \ref{tab:result_adv_example} shows typical examples, where the top two rows show an example for SEC, and the bottom two rows show an example for GED.
For example, the second example in Table \ref{tab:result_adv_example} was generated by changing {\it this} to {\it that}.
Even though this example does not alter the meaning, the prediction was changed from Negative $\to$ Positive.
The generated adversarial texts for GED still contain grammatical error; however the model predicts that they are grammatically correct.
Thus, these two examples are adversarial texts.

Note that the previous methods, AdvT-Text and VAT-Text, hardly reconstruct such effective adversarial texts.
Thus, this is a clear advantage of our methods compared with the previous ones.

\section{Conclusion}

This paper discussed the interpretability of adversarial training based on adversarial perturbation that was applied to tasks in the NLP field.
Our proposal restricted the directions of perturbations toward the locations of existing words in the word embedding space.
We demonstrated that our methods can successfully generate reasonable adversarial texts and interpretable visualizations of perturbations in the input embedding space, which we believe will greatly help researchers analyze a model’s behavior.
In addition, we confirmed that our methods, iAdvT-Text and iVAT-Text, maintained or improved the state-of-the-art performance obtained by our baseline methods, AdvT-Text and VAT-Text, in well-studied sentiment classification (SEC), category classification (CAC), and grammatical error detection (GED) benchmark datasets.

\section*{Acknowledgments}
We thank four anonymous reviewers for their helpful comments.
We also thank Takeru Miyato who suggested that we reproduce the result of a previous work \cite{miyato2016adversarial} as well as Masahiro Kaneko and Tomoya Mizumoto who provided helpful comments.

% \appendix
%
% \section{\LaTeX{} and Word Style Files}\label{stylefiles}
%
% The \LaTeX{} and Word style files are available on the IJCAI--18
% website, {\tt http://www.ijcai-18.org/}.
% These style files implement the formatting instructions in this
% document.
%
% The \LaTeX{} files are {\tt ijcai18.sty} and {\tt ijcai18.tex}, and
% the Bib\TeX{} files are {\tt named.bst} and {\tt ijcai18.bib}. The
% \LaTeX{} style file is for version 2e of \LaTeX{}, and the Bib\TeX{}
% style file is for version 0.99c of Bib\TeX{} ({\em not} version
% 0.98i). The {\tt ijcai18.sty} file is the same as the {\tt
% ijcai07.sty} file used for IJCAI--07.
%
% The Microsoft Word style file consists of a single file, {\tt
% ijcai18.doc}. This template is the same as the one used for
% IJCAI--07.
%
% These Microsoft Word and \LaTeX{} files contain the source of the
% present document and may serve as a formatting sample.
%
% Further information on using these styles for the preparation of
% papers for IJCAI--18 can be obtained by contacting {\tt
% pcchair@ijcai-18.org}.

%% The file named.bst is a bibliography style file for BibTeX 0.99c
\bibliographystyle{named}
\bibliography{ijcai18}

\begin{thebibliography}{}

\bibitem[\protect\citeauthoryear{Belinkov and
  Bisk}{2018}]{Belinkov2017SyntheticAN}
Yonatan Belinkov and Yonatan Bisk.
\newblock Synthetic and natural noise both break neural machine translation.
\newblock In {\em ICLR}, 2018.

\bibitem[\protect\citeauthoryear{Bengio \bgroup \em et al.\egroup
  }{2000}]{Bengio2000ANP}
Yoshua Bengio, R{\'e}jean Ducharme, Pascal Vincent, and Christian Janvin.
\newblock A neural probabilistic language model.
\newblock {\em Journal of Machine Learning Research}, 3:1137--1155, 2000.

\bibitem[\protect\citeauthoryear{Caruana \bgroup \em et al.\egroup
  }{2000}]{Caruana2000OverfittingIN}
Rich Caruana, Steve Lawrence, and C.~Lee Giles.
\newblock Overfitting in neural nets: Backpropagation, conjugate gradient, and
  early stopping.
\newblock In {\em NIPS}, 2000.

\bibitem[\protect\citeauthoryear{Dai and Le}{2015}]{Dai2015SemisupervisedSL}
Andrew~M. Dai and Quoc~V. Le.
\newblock Semi-supervised sequence learning.
\newblock In {\em NIPS}, 2015.

\bibitem[\protect\citeauthoryear{Goodfellow \bgroup \em et al.\egroup
  }{2015}]{Goodfellow2014ExplainingAH}
Ian~J. Goodfellow, Jonathon Shlens, and Christian Szegedy.
\newblock Explaining and harnessing adversarial examples.
\newblock In {\em ICLR}, 2015.

\bibitem[\protect\citeauthoryear{Grave \bgroup \em et al.\egroup
  }{2017}]{Grave2017EfficientSA}
Edouard Grave, Armand Joulin, Moustapha Ciss{\'e}, David Grangier, and
  Herv{\'e} J{\'e}gou.
\newblock Efficient softmax approximation for gpus.
\newblock In {\em ICML}, 2017.

\bibitem[\protect\citeauthoryear{Hochreiter and
  Schmidhuber}{1997}]{hochreiter1997long}
Sepp Hochreiter and J{\"u}rgen Schmidhuber.
\newblock Long short-term memory.
\newblock {\em Neural computation}, 9(8):1735--1780, 1997.

\bibitem[\protect\citeauthoryear{Hosseini \bgroup \em et al.\egroup
  }{2017}]{Hosseini2017DeceivingGP}
Hossein Hosseini, Sreeram Kannan, Baosen Zhang, and Radha Poovendran.
\newblock Deceiving google's perspective api built for detecting toxic
  comments.
\newblock {\em CoRR}, abs/1702.08138, 2017.

\bibitem[\protect\citeauthoryear{Jia and Liang}{2017}]{Jia2017AdversarialEF}
Robin Jia and Percy Liang.
\newblock Adversarial examples for evaluating reading comprehension systems.
\newblock In {\em EMNLP}, 2017.

\bibitem[\protect\citeauthoryear{Johnson and
  Zhang}{2015}]{Johnson2015SemisupervisedCN}
Rie Johnson and Tong Zhang.
\newblock Semi-supervised convolutional neural networks for text categorization
  via region embedding.
\newblock In {\em NIPS}, volume~28, pages 919--927, 2015.

\bibitem[\protect\citeauthoryear{Johnson and
  Zhang}{2016}]{Johnson2016SupervisedAS}
Rie Johnson and Tong Zhang.
\newblock Supervised and semi-supervised text categorization using lstm for
  region embeddings.
\newblock In {\em ICML}, 2016.

\bibitem[\protect\citeauthoryear{Kaneko \bgroup \em et al.\egroup
  }{2017}]{kaneko2017grammatical}
Masahiro Kaneko, Yuya Sakaizawa, and Mamoru Komachi.
\newblock Grammatical error detection using error-and grammaticality-specific
  word embeddings.
\newblock In {\em IJCNLP}, volume~1, pages 40--48, 2017.

\bibitem[\protect\citeauthoryear{Kingma and Ba}{2014}]{Kingma2014AdamAM}
Diederik~P. Kingma and Jimmy Ba.
\newblock Adam: A method for stochastic optimization.
\newblock {\em CoRR}, abs/1412.6980, 2014.

\bibitem[\protect\citeauthoryear{Le and Mikolov}{2014}]{Le2014DistributedRO}
Quoc~V. Le and Tomas Mikolov.
\newblock Distributed representations of sentences and documents.
\newblock In {\em ICML}, 2014.

\bibitem[\protect\citeauthoryear{Lehmann \bgroup \em et al.\egroup
  }{2015}]{Lehmann2015DBpediaA}
Jens Lehmann, Robert Isele, Max Jakob, Anja Jentzsch, Dimitris Kontokostas,
  Pablo~N. Mendes, Sebastian Hellmann, Mohamed Morsey, Patrick van Kleef,
  S{\"o}ren Auer, and Christian Bizer.
\newblock Dbpedia - a large-scale, multilingual knowledge base extracted from
  wikipedia.
\newblock {\em Semantic Web}, 6:167--195, 2015.

\bibitem[\protect\citeauthoryear{Lewis \bgroup \em et al.\egroup
  }{2004}]{Lewis2004RCV1AN}
David~D. Lewis, Yiming Yang, Tony~G. Rose, and Fan Li.
\newblock Rcv1: A new benchmark collection for text categorization research.
\newblock {\em Journal of Machine Learning Research}, 5:361--397, 2004.

\bibitem[\protect\citeauthoryear{Maas \bgroup \em et al.\egroup
  }{2011}]{Maas2011LearningWV}
Andrew~L. Maas, Raymond~E. Daly, Peter~T. Pham, Dan Huang, Andrew~Y. Ng, and
  Christopher Potts.
\newblock Learning word vectors for sentiment analysis.
\newblock In {\em ACL}, 2011.

\bibitem[\protect\citeauthoryear{Miyato \bgroup \em et al.\egroup
  }{2016}]{Miyato2015DistributionalSW}
Takeru Miyato, Shin ichi Maeda, Masanori Koyama, Ken Nakae, and Shin Ishii.
\newblock Distributional smoothing with virtual adversarial training.
\newblock In {\em ICLR}, 2016.

\bibitem[\protect\citeauthoryear{Miyato \bgroup \em et al.\egroup
  }{2017}]{miyato2016adversarial}
Takeru Miyato, Andrew~M Dai, and Ian Goodfellow.
\newblock Adversarial training methods for semi-supervised text classification.
\newblock In {\em ICLR}, 2017.

\bibitem[\protect\citeauthoryear{Nagata and
  Nakatani}{2010}]{Nagata2010EvaluatingPO}
Ryo Nagata and Kazuhide Nakatani.
\newblock Evaluating performance of grammatical error detection to maximize
  learning effect.
\newblock In {\em COLING}, 2010.

\bibitem[\protect\citeauthoryear{Pang and Lee}{2005}]{Pang2005SeeingSE}
Bo~Pang and Lillian Lee.
\newblock Seeing stars: Exploiting class relationships for sentiment
  categorization with respect to rating scales.
\newblock In {\em ACL}, 2005.

\bibitem[\protect\citeauthoryear{Rei and
  Yannakoudakis}{2016}]{rei-yannakoudakis:2016:P16-1}
Marek Rei and Helen Yannakoudakis.
\newblock Compositional sequence labeling models for error detection in learner
  writing.
\newblock In {\em ACL}, pages 1181--1191, 2016.

\bibitem[\protect\citeauthoryear{Rozovskaya \bgroup \em et al.\egroup
  }{2014}]{Rozovskaya2014TheIS}
Alla Rozovskaya, Kai-Wei Chang, Mark Sammons, Dan Roth, and Nizar Habash.
\newblock The illinois-columbia system in the conll-2014 shared task.
\newblock In {\em CoNLL Shared Task}, 2014.

\bibitem[\protect\citeauthoryear{Samanta and Mehta}{2017}]{samanta2017towards}
Suranjana Samanta and Sameep Mehta.
\newblock Towards crafting text adversarial samples.
\newblock {\em arXiv preprint arXiv:1707.02812}, 2017.

\bibitem[\protect\citeauthoryear{Srivastava \bgroup \em et al.\egroup
  }{2014}]{Srivastava2014DropoutAS}
Nitish Srivastava, Geoffrey~E. Hinton, Alex Krizhevsky, Ilya Sutskever, and
  Ruslan Salakhutdinov.
\newblock Dropout: a simple way to prevent neural networks from overfitting.
\newblock {\em Journal of Machine Learning Research}, 15:1929--1958, 2014.

\bibitem[\protect\citeauthoryear{Szegedy \bgroup \em et al.\egroup
  }{2014}]{Szegedy2013IntriguingPO}
Christian Szegedy, Wojciech Zaremba, Ilya Sutskever, Joan Bruna, Dumitru Erhan,
  Ian~J. Goodfellow, and Rob Fergus.
\newblock Intriguing properties of neural networks.
\newblock In {\em ICLR}, 2014.

\bibitem[\protect\citeauthoryear{Tokui \bgroup \em et al.\egroup
  }{2015}]{chainer_learningsys2015}
Seiya Tokui, Kenta Oono, Shohei Hido, and Justin Clayton.
\newblock Chainer: a next-generation open source framework for deep learning.
\newblock In {\em Proceedings of Workshop on Machine Learning Systems
  (LearningSys) in The Twenty-ninth Annual Conference on Neural Information
  Processing Systems (NIPS)}, 2015.

\bibitem[\protect\citeauthoryear{Yannakoudakis \bgroup \em et al.\egroup
  }{2011}]{yannakoudakis2011new}
Helen Yannakoudakis, Ted Briscoe, and Ben Medlock.
\newblock A new dataset and method for automatically grading esol texts.
\newblock In {\em ACL}, pages 180--189. ACL, 2011.

\end{thebibliography}

\end{document}